\documentclass[10pt,twocolumn,letterpaper]{article}

\usepackage{cvpr}              

\usepackage{graphicx}
\usepackage{amsmath}
\usepackage{amssymb}
\usepackage{booktabs}

\usepackage[square,numbers]{natbib}
\usepackage[pagebackref,breaklinks,colorlinks]{hyperref}

\def\cvprPaperID{00000} 
\def\confYear{2022}
\begin{document}

\title{It is Okay to Not Be Okay: Overcoming Emotional Bias in Affective Image Captioning by Contrastive Data Collection}

\author{Youssef Mohamed, Faizan Farooq Khan, Kilichbek Haydarov,  Mohamed Elhoseiny\\
{ King Abdullah University of Science and Technology (KAUST)} \\ 
\centerline{\small\texttt{\{youssef.mohamed,faizan.khan,kilichbek.haydarov,mohamed.elhoseiny\}@kaust.edu.sa}} \\
}

\makeatletter
\let\@oldmaketitle\@maketitle
\renewcommand{\@maketitle}{\@oldmaketitle
\myfigure\bigskip}
\makeatother
\newcommand\myfigure{%
\vspace{-7mm}
  \makebox[0pt]{\hspace{17.5cm}\includegraphics[width=1\linewidth]{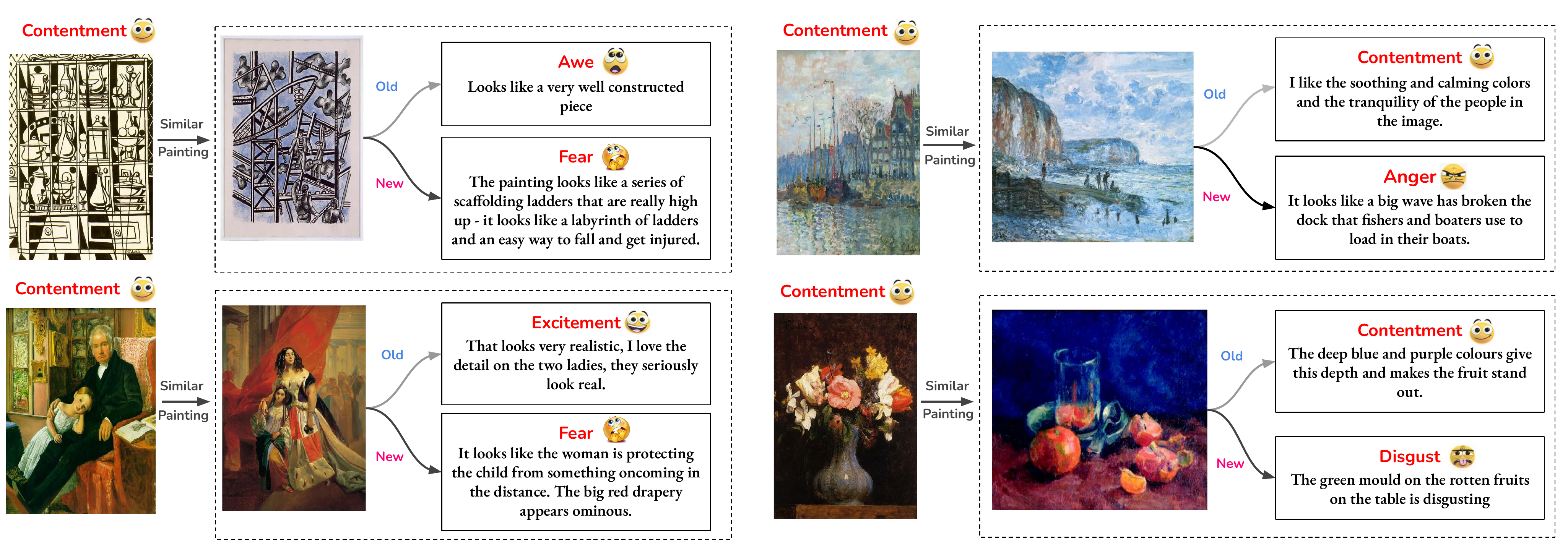}}
  \\
  \refstepcounter{figure}\normalfont{Figure~\thefigure:  \textbf{Examples from the contrastively collected dataset.} On the left side of each example is the query painting with its most common emotion on top of it. The right side shows a similar painting, based on the VGG feature map, which evokes the opposite emotion. We show the old utterance of the selected image and the new utterance to highlight the increased attention to details. Despite of paired paintings having very similar styles, the triggered emotions and utterances are very different.}
  \label{fig:teaser}
}
\maketitle
\begin{abstract}
Datasets that capture the connection between vision, language, and affection are limited, causing a lack of understanding of the emotional aspect of human intelligence. As a step in this direction, the ArtEmis dataset was recently introduced as a large-scale dataset of emotional reactions to images along with language explanations of these chosen emotions. We observed a significant emotional bias towards instance-rich emotions, 
making trained neural speakers less accurate in describing under-represented emotions. We show that collecting new data, in the same way, is not effective in mitigating this emotional bias. To remedy this problem, we propose a contrastive data collection approach to balance ArtEmis with a new complementary dataset such that a pair of similar images have contrasting emotions (one positive and one negative). We collected 260,533 instances using the proposed method, we combine them with ArtEmis, creating a second iteration of the dataset. The new combined dataset, dubbed {ArtEmis v2.0}, has a balanced distribution of emotions with explanations revealing more fine details in the associated painting. Our experiments show that neural speakers trained on the new dataset improve CIDEr and METEOR evaluation metrics by 20\% and 7\%, respectively, compared to the biased dataset. Finally, we also show that the performance per emotion of neural speakers is improved across all the emotion categories, significantly on under-represented emotions. The collected dataset and code are available at \url{https://artemisdataset-v2.org}.
\end{abstract}

\section{Introduction}

Emotional experiences stimulated by sensory information lie at the heart of human nature. They provide a window into rich yet less understood aspects of human intelligence. Emotions play a central role in determining humans' internal state, and subsequently, their behavior. Thus, studying emotional experiences and their expression is essential in understanding human behavior. Emotions are heavily influenced by external stimuli, especially vision and language. So, it is essential to have affective datasets that capture different modalities to study the relationship between sensory information and emotions. These datasets enable machines to comprehend emotions better and eventually increase social acceptance of AI applications, especially applications which interact with humans.

Several affective datasets connecting emotions to sensory information have been proposed. Most notably, GoEmotions dataset \citep{demszky2020goemotions} captured the underlying emotions behind Reddit comments. Its size was large enough to train deep learning models bringing machines a step closer to understanding emotions. However, a major drawback of GoEmotions and similar datasets \cite{cowen2017self, cowen2019mapping, cowen2020music, cowen2021sixteen} is that they attribute emotional experiences to a single stimulus.

Multi-modal datasets such as MS-COCO captions \cite{lin2014microsoft}, and VQA \cite{antol2015vqa} revolutionized the AI field and allowed machines to go beyond simple text/image problems towards  complex visual-language understanding tasks such as visual question answering and image captioning . These seminal datasets brought machines a significant step closer to human-level intelligence. \citet{achlioptas2021artemis} recognized that affective modeling needed similar multi-modal datasets to better understand emotions and how they are constructed. They introduced ArtEmis dataset that connects emotions, visual art, and language by collecting affective language explanations on visual artworks from the WikiArt dataset\cite{wikiartref}.
\paragraph{Dataset Bias}
\citet{plous2003psychology} suggests that biases and prejudices are integral to humans' evolution. He views biases as a method to optimize brain functions without the need for costly human attention. He claims that biases are created and modified by the human environment and experiences. That is why biases are extremely difficult to eliminate, but it is only possible to minimize their impact. Naturally, humans label datasets, inevitably introducing their biases in the collected data. These biases can sometimes be mild, but they can also be very problematic, especially in ethical judgment and applications that interact with humans \cite{torralba2011unbiased,biasexample1,kausel,ponce2006dataset}. \citet{biasexample1} showed that people ethically condemn certain behaviors based on a bad outcome, even though the outcome is determined randomly. 

Humans are usually capable of recognizing biases when they cause more harm than good. However, machine learning models do not have a similar ability to detect and reason about biases. Therefore, if models learn from a biased dataset, they will make biased decisions. Consequently, reducing biases from datasets is crucial in increasing acceptance and trust in machine learning models. It is equally important to detect biases in datasets, especially in affective datasets, used to train models that emulate human affect or interact directly with them. 

\citet{goyal2017makingv} identified a bias in VQA dataset \cite{antol2015vqa} making models trained on the dataset not rely properly on the visual modality and depend only on the language modality. This bias was introduced during the data collection process, and later this was observed in unexpected results from deep learning models trained on this dataset. The biased VQA1.0 had a fundamental issue with the distribution of answers to a given question, adversely affecting the trained models. Detecting and explaining this anomaly was not trivial because the distribution of VQA1.0 was biased. Consequently, the test set was biased, and despite the evaluation, metrics were high, giving the illusion of properly trained models; these models are not suitable for practical scenarios since the test set is not representative of the real world.

\begin{figure*}[t]
\centering
	\begin{subfigure}[b]{0.52\linewidth}
		\centering
		\includegraphics[width=1.\textwidth]{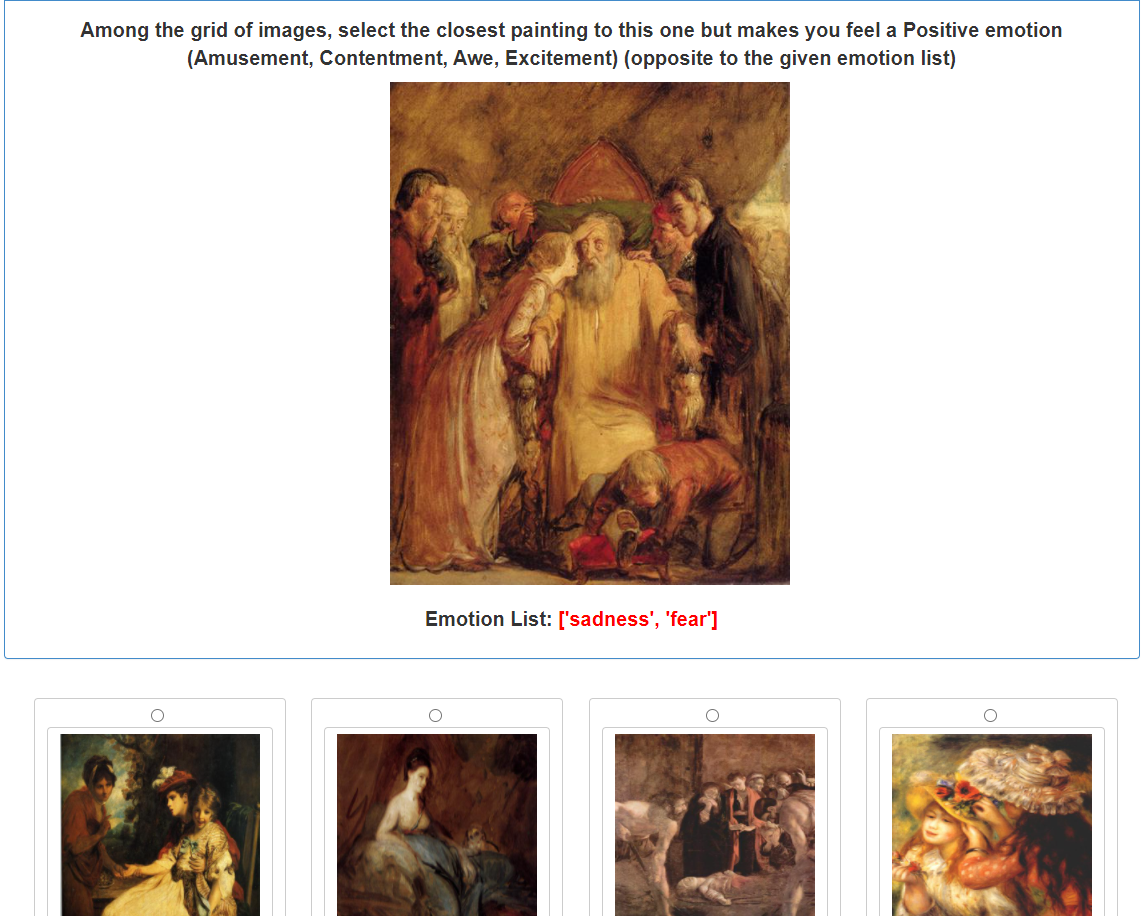}
		\caption{Most similar paintings selection (24, only 4 shown )}
		\label{fig:interface1}
	\end{subfigure}
    	\begin{subfigure}[b]{0.43\linewidth}
    		\centering
    		\begin{subfigure}[b]{1.\linewidth}
    		    \centering
    		    \scalebox{0.8}{
        		    \fbox{\includegraphics[width=1.\textwidth]{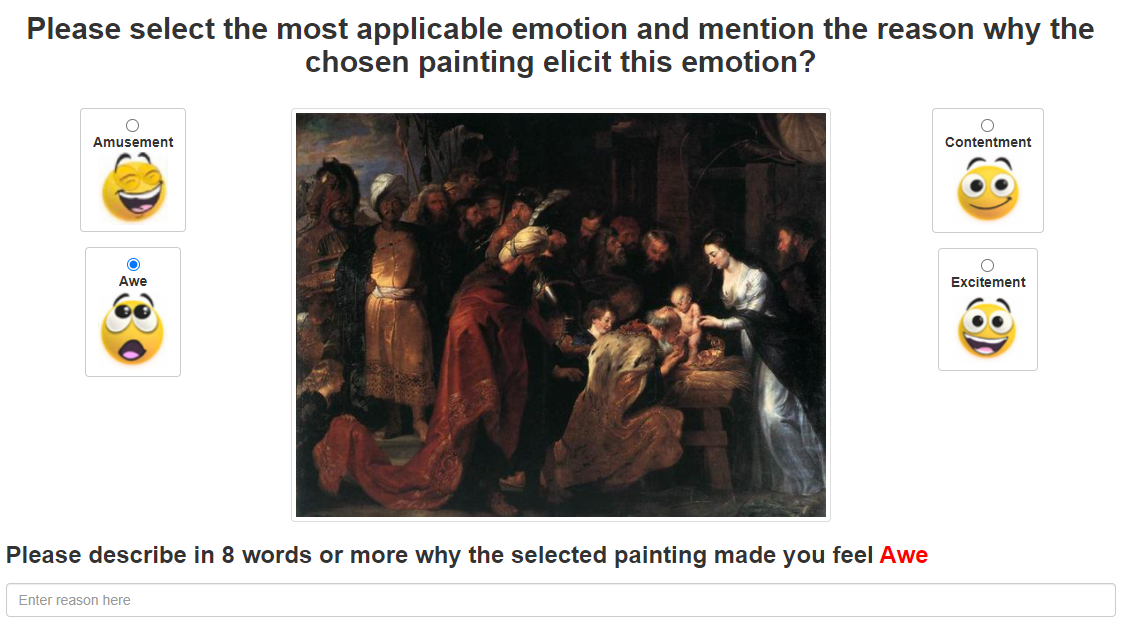}}
    		    }
    		    \caption{Positive emotion explanation interface.}
    		    \label{fig:interface2}
    		\end{subfigure}
    		\begin{subfigure}[b]{1.\linewidth}
    		    \centering
    		    \vspace{3mm}
    		    \scalebox{0.8}{
        		    \fbox{\includegraphics[width=1.\textwidth]{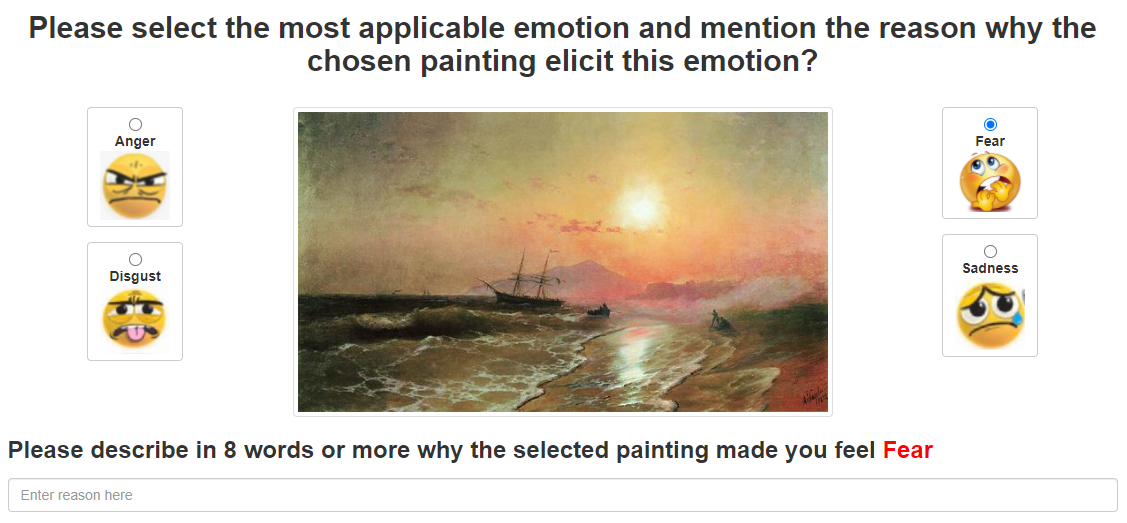}}
    		    }
    		    \caption{Negative emotion explanation interface.}
    		    \label{fig:interface3}
    		\end{subfigure}
    	\end{subfigure}
  \caption{\textbf{The contrastive data collection interface.} The left interface shows a painting and 24 of similar paintings from which the user selects the closest painting that elicits opposite emotions to the original one. A \emph{No Image Available} option can also be chosen. Depending on the sentiment of the original painting the user will either see the positive or negative emotion captioning interface (right interface).} 
\label{fig:interface}
\end{figure*}

Motivated by this, we observed that ArtEmis had a discrepancy in the results of trained neural speakers where a naive nearest neighbor model performed abnormally well. The main cause of this is an unbalanced distribution of emotions and generic captions. The unbalanced distribution is caused by a tendency of humans to feel positive about paintings. ArtEmis had 62\% of its captions labeled by a positive emotion compared to 26\% as negative, and the rest is something else. On the other hand, generic captions do not mention specific details about the described painting leading to less diversity among paintings with a similar style. For example, the two paintings in Fig. \ref{fig:teaser} in the bottom right corner have a similar style. In ArtEmis, the old caption describes the colors giving an overall contentment feeling and thus can match any caption from the neighborhood. On the other hand, the new caption recognizes the green patches as mold and thus elicits feelings of disgust. This detailed caption is very specific to this painting and can not be used for any neighboring painting. This diversity and attention to detail is something that ArtEmis lacks, and we attempt to solve this issue by collecting a complementary dataset using the interface we developed in Fig \ref{fig:interface}. We collect the complementary data in a contrastive fashion carefully designed to alleviate the bias in ArtEmis. Combining our complementary dataset with ArtEmis, we get a more balanced distribution of emotional labels where positive and negative emotions account for 47\% and 45\% of the dataset. We also show through experiments with neural speakers the superiority of this contrastive data collection method compared to expanding the size of ArtEmis using regular collection.

\vspace{0.2cm}
\noindent \textbf{Contributions.} 
 \begin{itemize}
    \item We show that the data collection process of the ArtEmis dataset results in an unbalanced distribution of emotions and generic captions, adversely affecting the quality of trained neural speakers. We also collect a complementary dataset using a \emph{Contrastive Data Collection} method, which alleviates the problems in ArtEmis
    \item We show that the captions in the combined dataset are more representative of Semantic Space Theory fine-grained emotions as studied in GoEMotions~\cite{demszky2020goemotions}.
     \item We trained multiple neural speakers to reflect the advantages of using contrastive data collection to complement ArtEmis. We show that speakers trained on our combined dataset significantly outperform speakers trained on ArtEmis in several aspects.
 \end{itemize}

\section{Related Work}

Several attempts to collect affective datasets led to datasets that help understand the connection between sensory information and emotions. Cowen et al. \cite{cowen2017self, cowen2019mapping, cowen2020music, cowen2021sixteen} collected emotional experiences induced as a result of different sensory information. However, these datasets are small in size, which limits their benefits in training deep learning models. GoEmotions dataset \cite{demszky2020goemotions} managed to capture the relation between emotional experiences and Reddit comments. Its size is also large enough to train text to emotion deep learning models and achieve good results. Nonetheless, all these datasets attribute emotional experiences to a single stimulus. On the other hand, ArtEmis \cite{achlioptas2021artemis} was collected, the first multi-modal dataset that captures the interplay of emotions, visual art, and language.

\begin{figure*}[t]
\begin{center}
    \scalebox{0.8}{
        \includegraphics[width=1\linewidth]{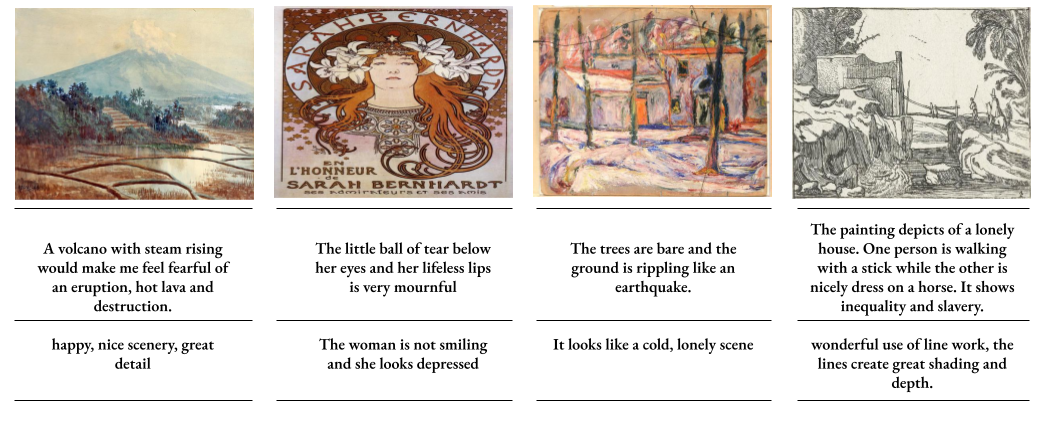}
    }
    \vspace{-3mm}
  \caption{\textbf{First Row:} new utterances collected using contrastive approach. Notice the difference in the image details referenced in each set. \textbf{Second Row:} original utterances collected from ArtEmis dataset. }
\label{fig:new_vs_old}
\end{center}
\end{figure*}

ArtEmis \cite{achlioptas2021artemis} dataset uniquely captures the complex relationship between visual stimuli, emotions as well as language. Captioning datasets, such as \cite{lin2014microsoft,young2014image, krishna2017visual,kazemzadeh2014referitgame, sharma2018conceptual, mao2016generation, pont2020connecting}, contain only factual explanations for a given image. In contrast, ArtEmis welcomes subjectivity, producing diverse utterances enriched with abstract concepts and emotional states.
For example, 
the volcano in the leftmost painting in Fig.\ref{fig:new_vs_old} can evoke awe emotion due to its majesty or fear emotion when a viewer think of lava and destruction caused by an eruption. Whilst a factual caption would describe it as just a volcano.
These imaginative and symbolic aspects of ArtEmis Dataset are crucial in producing more human-like AI. We mainly consider ArtEmis dataset in this work because it is the only dataset that captures the interplay among vision, language, and emotions to our best knowledge. 

The proliferation of large-scale captioning datasets allowed for the development of many deep neural network based captioning methods \cite{lu2018neural,mao2016generation,vedantam2017context,yu2016modeling,nagaraja2016modeling,yu2018mattnet}. We are interested in applying captioning models in this paper, so we chose state-of-the-art standard models, in particular, \citet{xu2015show} where they used an LSTM model with attention and \citet{cornia2020meshed} where they adopted a transformer for the captioning task. 
We show that training neural speakers with our complementary dataset implicitly leads to better performance and higher caption specificity.

\paragraph{Connection to Emotion Theories.} We view ArtEmis as a unique dataset that can act as a bridge between different emotion theories, in particular the well established \emph{Theory of Constructed Emotions} \cite{barrett2006solving, barrett2007mice, barrett2017emotions, barrett2017theory} and the recently proposed {\emph{Semantic Space Theory of Emotions}} \cite{cowen2020semantic}. The language explanation of the emotional experience evoked from looking at an artwork is better viewed in the context of the theory of constructed emotions. This theory considers language the main way to communicate affective experiences while emotion categories such as anger are subjective and differ from one person to another, making them less effective. On the other hand, semantic space theory argues that emotion categories/labels are a better method to communicate emotional experiences. They identified a wide range of emotional categories ranging from 18 to 25 depending on the stimulus. Moreover, they show that these categories have smooth boundaries between them contrary to the belief that emotional categories are discrete \cite{darwin2015expression, tomkins1984affect,boyle1984reliability,izard1993stability, boyle2015criteria}. ArtEmis blends the two theories by providing both language and emotional categories. It provides an opportunity to develop a unified emotion theory that uses language and emotional categories to explain affective experiences. However, the bias we detected in ArtEmis, greatly limits its ability since its emotional distribution is unbalanced, and the captions have low specificity to their paintings.

\section{Motivation}
\label{databias}
\cite{achlioptas2021artemis} reported the results of training multiple neural speaker architectures on the ArtEmis dataset. Among these models, they used a naive Nearest Neighbor (NN) Retrieval model, which identifies the closest neighbors to a test painting from the training paintings and randomly selects an utterance from this neighborhood. They also trained Meshed Memory Transformers \cite{cornia2020meshed} as well as ``Show, Attend and Tell''  model \cite{xu2015show}. Nearest neighbor model achieved scores of 0.102 and 0.210 in METEOR and ROUGE metrics, respectively, compared to 0.140 and 0.280 for meshed memory transformer and 0.142 and 0.297 for show, attend and tell. The NN results are unexpectedly good compared to the other models. This high performance can be attributed to a lack of diversity in the neighborhood of each painting. This greatly helps the nearest-neighbor model since the utterances in the local context of each sample are similar. On the downside, neural speakers trained using ArtEmis exhibit a lack of specificity to the target image and tend to produce generic captions.

Annotations in ArtEmis are highly subjective, and some paintings elicit opposite emotions depending on the annotator. However, for most of the dataset, paintings do not elicit opposite sentiment emotions, and the neighboring paintings have a similar emotional sentiment. We follow \cite{achlioptas2021artemis} and define positive sentiment emotion as one of Contentment, Awe, Excitement or Amusement; while a negative sentiment emotion as one of Anger, Disgust, Fear, or Sadness. We identified \emph{33987} paintings with single sentiment emotions \textit{i.e.} they do not evoke opposite emotions. For each single sentiment paintings, the ratio of paintings in its \(K\) neighbor, which has a similar sentiment, is {40\%} over \(K\) values ranging from 2 - 10. This number is very high considering the subjective nature of ArtEmis, and it is the main reason for the Nearest Neighbor model's high performance. 

Further inspection also revealed many instances where the utterance for the selected emotion is very generic such as those shown in the bottom row in Fig. \ref{fig:new_vs_old}. These utterances do not mention any details in the paintings despite the paintings expressing real objects and providing a good context that the utterance can better utilize. This lack of fine details and the lack of contrastive emotional neighborhood around many paintings negatively affect the performance of neural speakers since the captions are generic and similar to each other for similar emotions. 

We propose to collect a complementary dataset to alleviate the problems above. Our goal is to introduce more \emph{diversity} in the neighborhood of single sentiment paintings. We also aim to enhance the dataset's quality of utterances, making them more grounded/specific to the paintings. By doing so, trained neural speakers on the new augmented dataset will learn to attend more to the fine details of every painting in order to have a quality utterance that explains the elicited emotion properly.

\section{Contrastive Dataset}
\label{methodology}
\subsection{Data Collection Interface}
In order to introduce more emotional diversity, we collected the new complementary dataset in a contrastive manner.
We defined an emotional score for each painting as 
\(
score_i = \frac{pos_i - neg_i}{N_i}
\)
where $pos_i$ and $neg_i$ are the number of positive and negative emotions respectively, while $N_i$ is the total number of emotions associated with the $i$-th painting. We identified emotionally biased paintings with an absolute emotional score greater than $0.3$. 
Then for each of those painting, we retrieved the nearest ${100}$ neighbors. The nearest neighbors were identified based on high-level semantic features extracted from layer \emph{fc7} of VGG16 Network \cite{vgg}. Out of these ${100}$, we chose ${24}$ paintings. The first ${12}$ were the 12 nearest neighbors, while the last ${12}$ had the highest emotional score among the remaining paintings with the same sentiment as the query one. 
Hence, the 24 paintings contains visually similar paintings and evoke similar emotions to the query painting. By design, this encourages the participants to pay more attention to contrastive details that construct opposite emotional experiences; see Fig.~\ref{fig:teaser}. 

After obtaining the complete list of neighbors for the identified paintings, we launched collection experiments on Amazon Mechanical Turk (AMT). We have two tasks: in the first one, we used the interface shown in Fig.~\ref{fig:interface1}. Given a random query painting with its list of emotions,  we ask AMT workers to select the most similar painting from its 24 visual nearest neighbors, which evokes an \textbf{opposite} emotion. If turkers do not find a proper painting, we allow them to choose a \emph{No Image Available} option, to avoid imposing an emotional bias on them.
In the second task, once the painting is selected, we ask annotators to specify the primary emotion they felt by observing the selected painting using a similar interface used by \cite{achlioptas2021artemis}.
The interfaces, shown in Fig. \ref{fig:interface2} and \ref{fig:interface3}, ask the turker to select an emotion and mention the reason why. We report statistics about the data collection task in Section \ref{datacollectresults}. We thoroughly reviewed all the collected data to ensure its high quality.

\subsection{Collected Data Statistics}
\label{datacollectresults}
In total, we identified \emph{52933} emotionally biased paintings. We collected a total of $\textbf{260,533}$ instances, allowing at least five submissions per painting. Out of the collected data, only 7752 had the \emph{No Image Available} option selected, accounting for $3\%$ of our dataset. This small number reveals that turkers, upon closer inspection, can extract details from most paintings that may elicit contradicting emotions. To measure the diversity of the dataset, we calculated the entropy of emotions in the $K$ visual neighbors for every painting. For $K=20$, ArtEmis has an entropy of 0.805 while combining the complementary dataset with ArtEmis resulted in entropy of 0.855, a total of 6\% increase, reflecting an increase in the local diversity of every painting.
We name our complementary data as $Contrastive$ and combine it with a random subset from $ArtEmis$ resulting in a dataset of the same size. We name this dataset $Combined$ and mainly compare it to $ArtEmis$ highlighting the advantages of using contrastive data collection. To guarantee fairness, we make sure all the datasets have similar sizes.

\subsection{Qualitative analysis}
\label{qualitative}
Samples from $Contrastive$ are shown in Fig. \ref{fig:teaser}. The query image is shown on the left, and the selected nearest painting which evokes opposite emotions on the right, along with an old utterance and a newly collected one. The original explanations of the selected painting reflect their shallowness and the lack of attention to fine details. The constraint placed on the turkers to choose a painting evoking opposite emotion made them focus on the details of each image and perform more emotional associations. For example, in the bottom right example, the query painting evoked contentment emotion, and the nearest painting originally evoked contentment as well. However, by observing the painting, an annotator feels disgusted because of the green tone, which resembles mold. 

We highlight the increase in attention to fine details in Fig. \ref{fig:new_vs_old}, where we showcase three paintings with different art styles. The bottom and top rows show old and new utterances, respectively, contrasting the improved depth of details. For example, the new utterance of the painting on the left describes the volcano and how the annotator is afraid that it might erupt; in contrast, the caption from ArtEmis is very generic and can suit any painting. In the middle painting, despite the old utterance mentioning the woman, it pales in comparison to the details mentioned in the new utterance. Finally, we see the same comparison in the right painting, where the new utterance mentions unique similes.

\begin{figure}[t]
	\centering
	\begin{subfigure}[b]{1\linewidth}
	\centering
	\scalebox{0.8}{
		\includegraphics[width=1.\textwidth, height=0.7\textwidth]{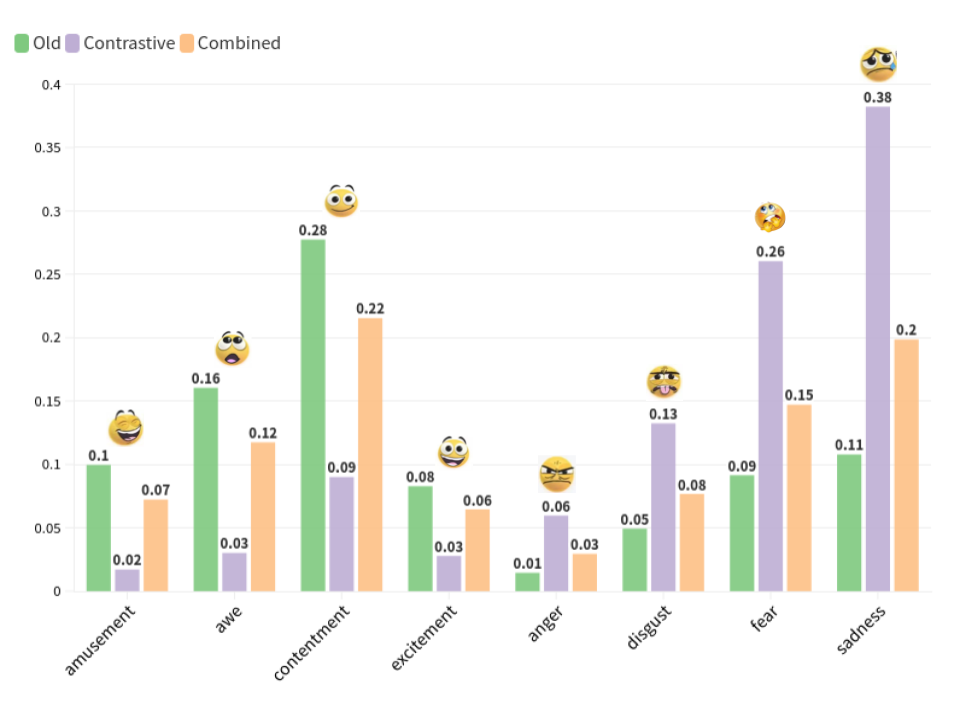}
		}
		\caption{There is an increase in the count of the negative emotions but some emotions are still under-annotated.}
    	\label{fig:emo_dist}
	\end{subfigure}
    \begin{subfigure}[b]{1\linewidth}
	    \centering
	    \scalebox{0.6}{
		    \includegraphics[width=1.\textwidth, height=0.6\textwidth]{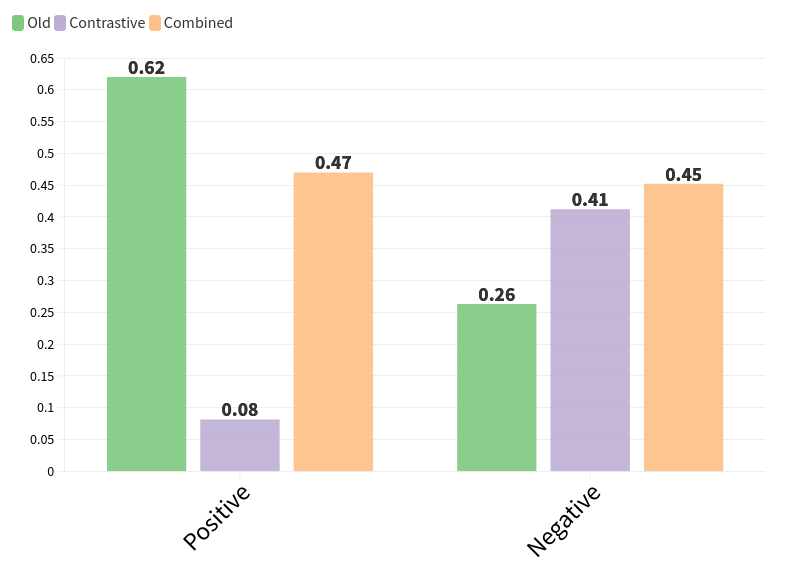}
    	}
	\caption{The semantic distribution of emotions is much more balanced when combining the datasets.}
    \label{fig:emo_dista}
	\end{subfigure}
	\caption{\textbf{Combining the complementary dataset with ArtEmis results in a more balanced distribution of emotions.} Note how the complementary dataset naturally gets more annotations for under-annotated emotions in ArtEmis.}
\end{figure}

\subsection{Quantitative analysis}
\label{quantitative}
\paragraph{Emotion Distribution}
Our setup for the data collection inherently balances the emotions in ArtEmis. This is evident from the emotion distribution shown in Fig. \ref{fig:emo_dist}. Originally, ArtEmis dataset had \textbf{62\%} positive emotions compared to only \textbf{26\%} negative emotions (the rest is something else), making most query paintings have a positive sentiment. Consequently, the turkers are limited to negative emotions or ``No Image'' option for most tasks which ultimately balances the emotions distribution. As a result, $Combined$ dataset has a more balanced distribution, with \textbf{47\%} of the samples being positive and \textbf{45\%} being negative. 
Note that here we contrast the sentiment of the utterances, not the exact emotions. This is why the balance is evident in the level of sentiment distribution.

\begin{figure}[t]
	\centering
	\scalebox{0.45}{
    	\includegraphics[width=1.\textwidth]{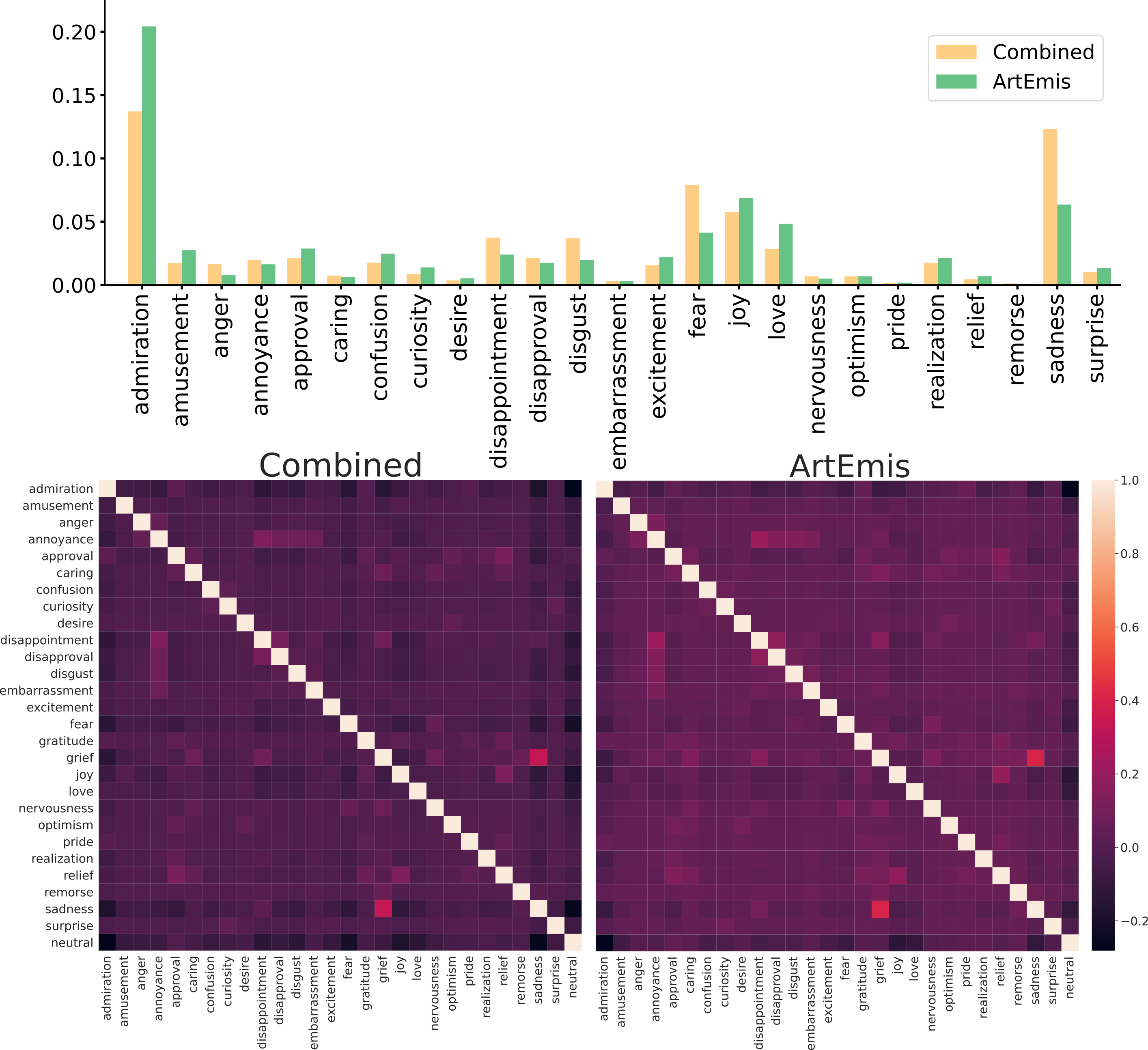}
	}
	\caption{\textbf{Semantic Space Theory  Fine-grained Emotion Analysis.} Top: we plot the histogram over the extended emotion set from GoEmotion. Note how the distribution is more balanced in $Combined$. Bottom: we show the correlation of the emotions in $Combined$ and $ArtEmis$. The darker off-diagonal patches mean $Combined$ has less correlation between different emotions, more  distinctively representing them.}
	\label{fig:go_emo}
\end{figure}

We further analyze the emotional distribution of the combined dataset by extending the fine-grained emotion set according to the semantic space theory~\cite{cowen2020semantic}. We fine tune a RoBERTa~\cite{liu2019roberta} language model on the GoEmotions dataset. In in Fig. \ref{fig:go_emo}, we then use this model to predict the extended emotion set of both $Combined$ and $ArtEmis$. We plot the histogram of the emotion responses and the  Pearson correlation between all pairs of emotion types. The figure shows that the combined dataset is more representative of the Semantic Space theory emotions compared to ArtEmis evident from the off-diagonal correlations being darker in the combined dataset compared to ArtEmis. For instance, the fear and disappointment emotions have lower correlation in the combined heat-map revealing that the captions in the combined dataset \emph{distinctively} express these emotions.

\paragraph{Parts-Of-Speech Analysis}
\label{sec:ling}
We compare the linguistic structure of the captions in $Contrastive$, $Combined$, and $ArtEmis$ in Table \ref{tab:lingstats}. The linguistic structure of the three datasets is similar, with minor differences. This similarity shows that the $Combined$ dataset does not have an unfair linguistic advantage, such as fewer words per caption.
One observation worth noting is the increase in the verbs per caption, which may reflect a rise in the number of associations. Associations are used primarily to relate different parts of a painting; this can be interpreted as more attention to the painting's details. 
These results suggest that the performance gain we achieve in Section \ref{sec:dataset_comp} can be mostly attributed to the balanced emotion distribution as well as attention to fine details.

\begin{table}[t]
	\centering
	\scalebox{0.6}{
		\begin{tabular}{@{}lcccccc@{}}
			\toprule
			\multicolumn{1}{l}{Dataset} & Words & Nouns & Pronouns & Adjectives & Adpositions & Verbs \\
			\midrule
			\multicolumn{1}{l}{Contrastive} & 15.8 & 3.8 & 0.9 & 1.5 & 1.8 & \textbf{3.3} \\
			\multicolumn{1}{l}{ArtEmis}\cite{achlioptas2021artemis}     & 15.9 & \textbf{4.0} & 0.9 & 1.6 & 1.9 & 3.0 \\
			\multicolumn{1}{l}{Combined}    & 15.9 & 3.9 & 0.9 & 1.6 & 1.9 & 3.2 \\
			\bottomrule
		\end{tabular}
	}
	\caption{ \textbf{The richness of captions reported as the average linguistic units per individual captions.} Contrastive is our contrastively collected dataset, Combined is the union of ArtEmis and Contrastive.}
	\label{tab:lingstats}
\end{table}

\section{Experiments}

\noindent \textbf{Neural Speakers}
We follow \cite{achlioptas2021artemis} in training and evaluating different Neural Speakers (\emph{Affective} Image Captioning Models). The first is a naive model based on K-nearest neighbors. We identify the nearest neighbors based on high-level semantic features extracted from layer \emph{fc7} of VGG16 net. For inference, we retrieve the nearest $3$ neighbors from the training set, randomly selecting one caption.
We also train ``Show, Attend and Tell'' (SAT) \cite{xu2015show} which is based on LSTM~\cite{hochreiter1997long} and Meshed-Memory Transformers ($M^2$)~\cite{cornia2020meshed}, the state-of-the-art captioning model on MS-COCO dataset. In addition, we developed a modified version of $M^2$ transformer, which is better adapted for artworks. 
$M^2$ transformer uses object features as image representation which might not be suitable for paintings in ArtEmis since some artworks do not depict real objects (e.g., abstract paintings). Therefore, we propose to extract patch features from paintings by dividing the painting into $P\times P$ patches. We then extract high-level features from the last convolutional block of VGG16. We concatenate the patch features with object features to get a diverse set of representations. In our experiments, we set $P=4$. 
For fair comparisons between the different models, we downscaled the hidden sizes of the different models such that they have roughly similar time complexity. Each model except for the NN baseline takes 4-5 hours of training time on a single Nvidia V100 GPU.

\noindent \textbf{Training Sets.} We define three datasets: $Contrastive$ contains 260,533 samples we collected, $ArtEmis$ contains all samples from the original ArtEmis dataset~\cite{achlioptas2021artemis}, finally, $Combined$ is the union of $Contrastive$ and 260,533 random samples from ArtEmis. For fair comparisons, we randomly removed 65K captions from $Combined$, such that it has the same size as $ArtEmis$ \emph{i.e.} 455K. 

\noindent \textbf{Test sets.} For evaluation, we test on two sets. The first is a subset of size $10\%$ from $Combined$, while the second is a subset provided by \cite{achlioptas2021artemis} called $ArtEmis_{C40}$ which is not included in $ArtEmis$ but was collected in the same fashion. $ArtEmis_{C40}$ contains 703 paintings, each having at least 40 emotions and corresponding explanations.
We chose $ArtEmis_{C40}$ since it has more samples per painting, allowing for more subjectivity and diversity, and hence help more accurately measure the quality of the generated captions of a given emotion. 

We highlight the advantages of collecting data contrastively by reporting the results of three experiments. The first is a benchmark of the aforementioned neural speakers. Secondly, we compare the evaluation scores of Three SAT models trained on: $Combined$, $ArtEmis$, and $Contrastive$. Finally, we break down the results per emotion, showing significant gains in performance across all emotions, especially for underrepresented emotions.
\begin{table}[t]
    \centering
    \scalebox{0.7}{
		\begin{tabular}{lcccc}
		    \toprule 
			Metric & NN & \(M^2\) & modified \(M^2\) & SAT \\
			\midrule 
			BLEU-1  & 0.145 & 0.558   & 0.565 &\textbf{0.628} \\
			BLEU-2  & 0.040 & 0.338   & 0.339 & \textbf{0.385} \\
			BLEU-3  & 0.013 & 0.202   & 0.201 & \textbf{0.226} \\
			BLEU-4  & 0.005 & 0.123   & 0.123 & \textbf{0.137} \\
			METEOR  & 0.057 & 0.147   & 0.146 & \textbf{0.165} \\
			ROUGE-L & 0.124 & 0.307   & 0.309 & \textbf{0.339} \\
			CIDEr   & 0.048 & 0.091   & 0.096 & \textbf{0.103} \\
			\bottomrule 
		\end{tabular}
	}
    \caption{\textbf{Results of neural speakers trained on the Combined dataset}. We evaluate different neural speakers on the combined test set. SAT outperforms the other models, while the nearest neighbor (NN) model performance drops significantly compared to the model in \cite{achlioptas2021artemis}.}
	\label{tab:benchmark}
	\vspace{-2mm}
\end{table}

\begin{table}[t]
    \centering
    \scalebox{0.7}{
		\begin{tabular}{lcccc}
		    \toprule 
            Metric & $Combined$ & $ArtEmis$ & $Contrastive$ \\
			\midrule 
			BLEU-1  & \textbf{0.855 / 0.540} & 0.837 / 0.511 & 0.820 / 0.521\\
			BLEU-2  & \textbf{0.665 / 0.301} & 0.642 / 0.282 & 0.613 / 0.283\\
			BLEU-3  & \textbf{0.480 / 0.168} & 0.456 / 0.154 & 0.425 / 0.151\\
			BLEU-4  & \textbf{0.338 / 0.096} & 0.313 / 0.088 & 0.287 / 0.084\\
			METEOR  & \textbf{0.218 / 0.144} & 0.212 / 0.135 & 0.204 / 0.135\\
			ROUGE-L & \textbf{0.449 / 0.295} & 0.447 / 0.284 & 0.433 / 0.284\\
			CIDEr   & \textbf{0.086 / 0.111} & 0.076 / 0.091 & 0.080 / 0.093\\

			\bottomrule   
		\end{tabular}
	}
	\caption{\textbf{SAT performance trained on different training sets}. The four different SAT models are evaluated on $ArtEmis_{C40}$ test set. We report the results  in this table  averaged per caption (before /) and  per emotion (after /). Note that SAT model trained using $Combined$ outperforms all the other models.}
	\label{tab:c40}
		\vspace{-3mm}
\end{table}

\begin{figure*}[t]
	\centering
    \scalebox{0.9}{
	    \includegraphics[width=1\textwidth, height=0.55\textwidth]{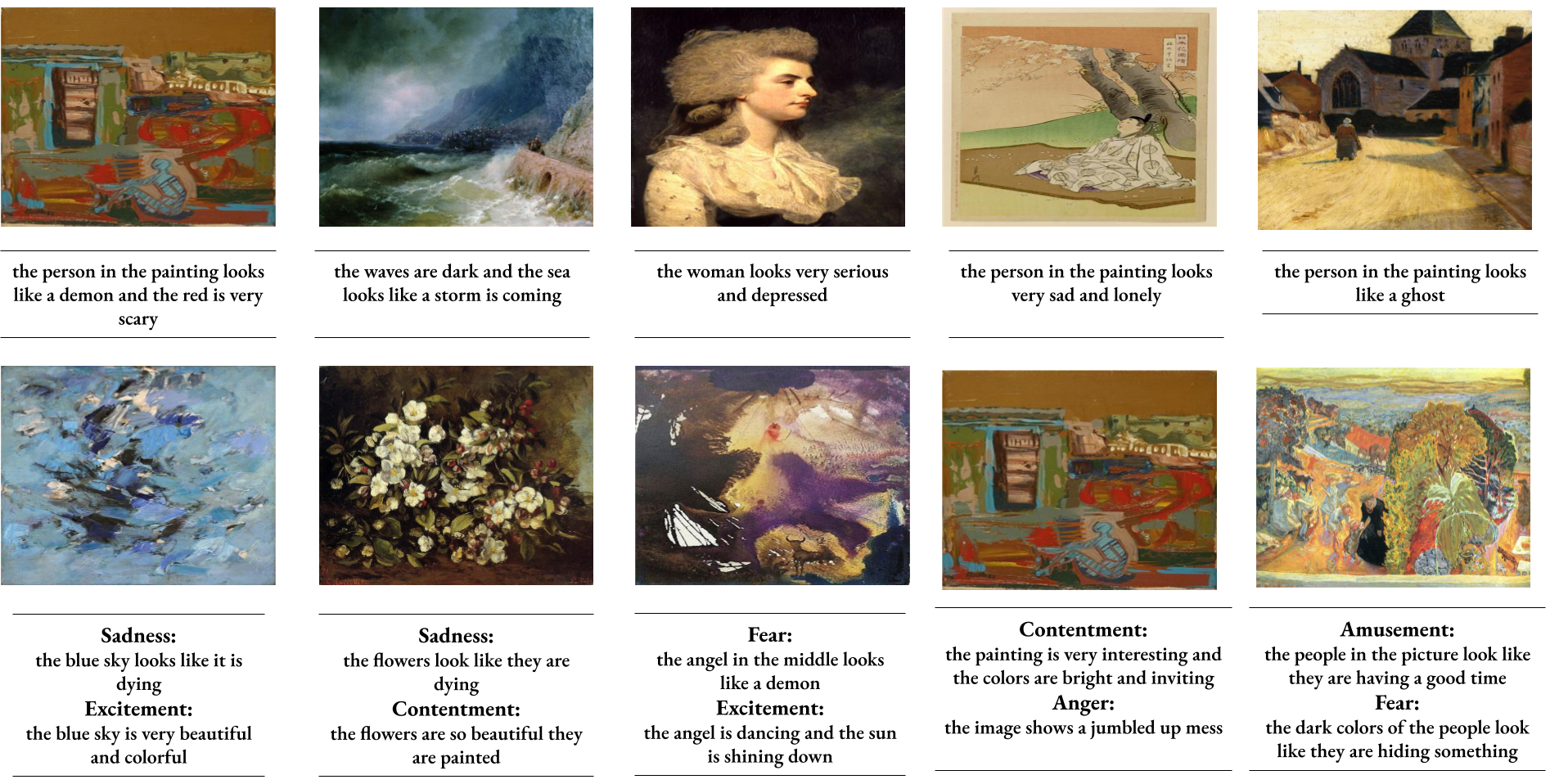} 
	}
	\caption{\textbf{First Row:} generations from SAT model. \textbf{Second Row:} generations from emotionally grounded SAT model. Both models are trained on $Combined$ dataset. Note the attention to specific details in each painting.}
	\label{fig:sample_gens}
	\vspace{-2mm}
\end{figure*}
\noindent \paragraph{Results.}
We use $Combined$ dataset and perform a similar benchmark to the one done in \cite{achlioptas2021artemis}. we report the metrics measured on the $Combined$ test set. We compare NN, SAT, and M2 (vanilla and modified versions). The results reported in Table \ref{tab:benchmark} show how the naive NN model suffers on this combined dataset. For example, for METEOR and ROUGE-L scores, the relative improvement between SAT over NN for ArtEmis~\cite{achlioptas2021artemis} was \textbf{28}\% and \textbf{29}\%, respectively. On the other hand, for $Combined$ dataset, the improvement becomes \textbf{65}\% and \textbf{63}\%. This drastic decrease in NN performance reflects the diversity introduced in the $Combined$ dataset, making it harder for the NN model to perform well. SAT is the best performing model on most evaluation metrics except for CIDEr. The modified \(M^2\) outperforms \(M^2\) slightly, supporting our claim that using only bounding boxes features is not suitable for paintings.
Due to its superiority, we use SAT in the next experiments to explore the advantages of the $Combined$ dataset.
\paragraph{Datasets Comparison.}
\label{sec:dataset_comp}
We evaluate four emotionally grounded SAT models trained on  $Combined$, $ArtEmis$, and $Contrastive$. To guarantee fair comparison, we use $ArtEmis_{C40}$ since it was collected by \cite{achlioptas2021artemis} and there are no newly collected samples included in it. The results are reported in Table \ref{tab:c40}. The model trained on $Combined$ performs the best. This suggests that using a balanced training set via adding contrastive data, significantly improves neural speakers. This can be attributed to the improved representation discriminativeness, due to training on visually similar images with opposite emotions. 
Notably, in spite of having half the size, the model trained on $Contrastive$ come close to the model trained on $ArtEmis$, indicating that captions collected contrastively have more specific descriptions of the paintings.

\begin{figure}[b!]
	\centering
    \vspace{-2mm}
    \scalebox{0.4}{
	    \includegraphics[width=1.\textwidth]{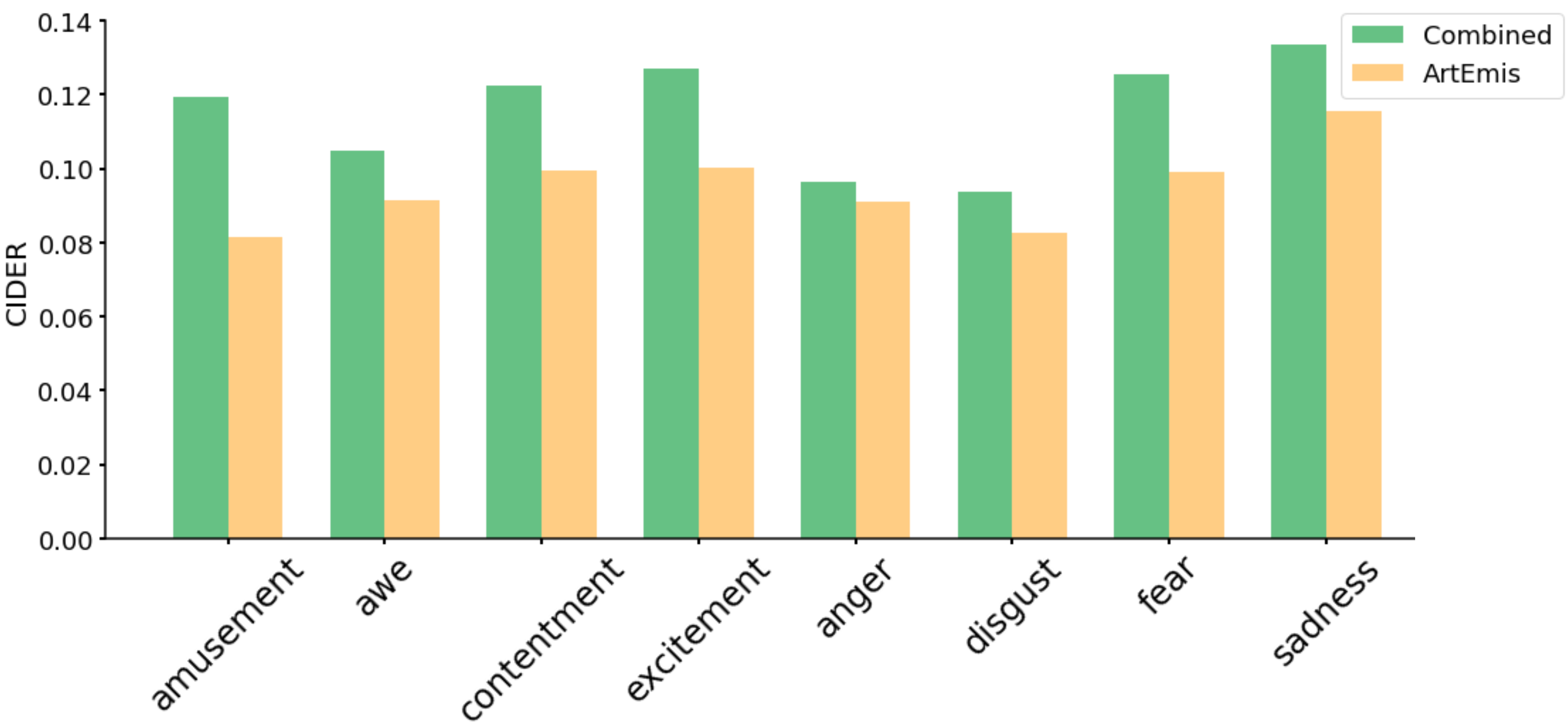} 
	}
	    \scalebox{0.4}{ 
	    \includegraphics[width=1.\textwidth]{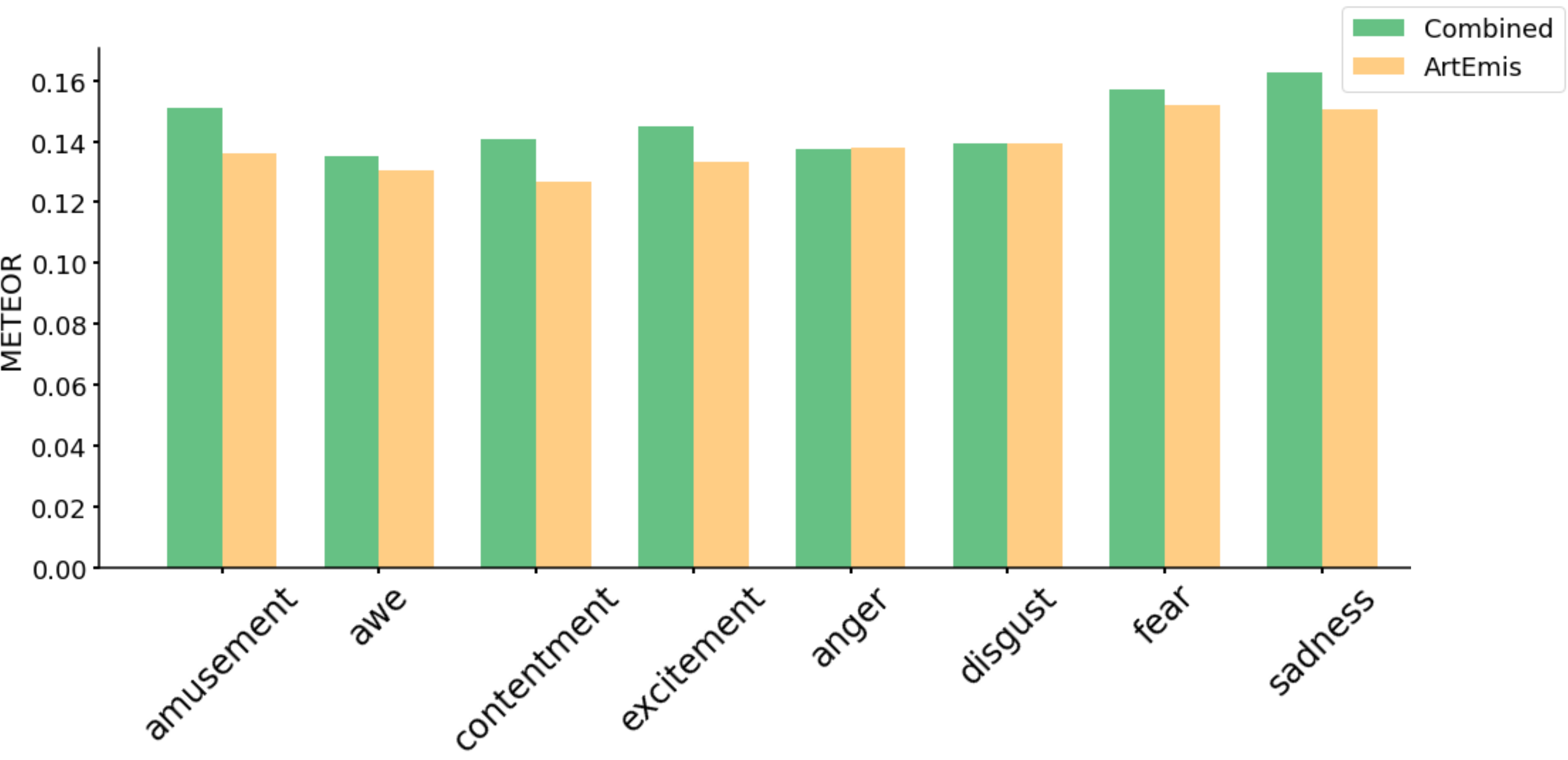}
	}
	    \vspace{-2mm}
	\caption{\textbf{Scores per grounding emotions.} SAT improves significantly when trained on $Combined$ versus ArtEMis on all emotions. CIDEr and METEOR improves by 20\% and 7\%, also note how the difference in performance is more significant for less frequent emotions.}
	\label{fig:score_emo}
\end{figure}

\vspace{-0.3cm}
\paragraph{Per Emotion Analysis.}
We deeply analyze the scores by averaging the results per emotion. We group the samples in $ArtEmis_{C40}$ by the emotions, then calculate the evaluation metrics per caption for each emotion separately; finally, we report the average scores per emotion. Fig. \ref{fig:score_emo} demonstrates the superiority of the model trained on $Combined$ on every emotion, especially for \emph{underrepresented} emotions which were less frequent in ArtEmis, such as amusement, excitement, and all the negative emotions. We report the average score per emotion in Table~\ref{tab:c40}, the performance gap increases drastically where CIDEr and METEOR improve by 20\% and 7\%, respectively.
We preview a sample of generations in Fig.~\ref{fig:sample_gens}. The top row shows generation from a non-emotionally grounded SAT model, while the second model generates captions based on input emotion. The generations are high quality and reflect details specific to the paintings. 
\vspace{-2mm}
\paragraph{Human Experiment.}  We also performed human evaluation on 100 randomly sampled paintings (5 responses each) with generations from SAT trained on $Combined$ versus $ArtEmis$; \textbf{73\%} favoured $Combined$ generations.
\section{Conclusion}
\label{conclusion}
This paper identifies bias problems in the task of Affective Image captioning, particularly unbalanced emotion distribution and generic captions, negatively affecting the quality of trained models. We introduced ArtEmis v2, a second iteration of ArtEmis to mitigate the emotional bias. We collected our data in a contrastive manner that by design balances ArtEmis and encourages annotators to pay extra attention to fine details. We analyzed the new combined dataset revealing its advantages, and conducted several experiments to show how neural speakers perform better when trained on the combined dataset. Please note that Affective datasets might have other biases that are yet to be addressed, including bias towards ethnic groups and minorities.
We have focused only on the emotional bias in this work, and we hope our contrastive data collection approach introduces a step towards mitigating other forms of biases in Affective Vision and Language datasets.

\noindent \textbf{Acknowledgments.} This work is supported by KAUST,  under Award No. BAS/1/1685-01-01.

{\small
\bibliography{egbib}
}

\end{document}